\documentclass[10pt,twocolumn,letterpaper]{article}

\usepackage[pagenumbers]{iccv} 

\usepackage{amsmath,amsfonts}
\usepackage{algpseudocode}
\usepackage{multirow}
\usepackage{cuted}

\usepackage{float}

\newcommand{\squishlist}{
   \begin{list}{$\bullet$}
    { \setlength{\itemsep}{0pt}      \setlength{\parsep}{3pt}
      \setlength{\topsep}{3pt}       \setlength{\partopsep}{0pt}
      \setlength{\leftmargin}{1.0em} \setlength{\labelwidth}{1em}
      \setlength{\labelsep}{0.5em} } }
      
\newcommand{\squishend}{
    \end{list}  }

\definecolor{iccvblue}{rgb}{0.21,0.49,0.74}
\usepackage[pagebackref,breaklinks,colorlinks,allcolors=iccvblue]{hyperref}

\def\paperID{2935} 
\def\confName{ICCV}
\def\confYear{2025}

\title{Transform-Dependent Adversarial Attacks}

\author{Yaoteng Tan, Zikui Cai, and M. Salman Asif\\
University of California Riverside\\
{\tt\small \{ytan082,zcai032,sasif\}@ucr.edu}
}

\begin{document}
\maketitle
\begin{abstract}
Deep networks are highly vulnerable to adversarial attacks, yet conventional attack methods utilize static adversarial perturbations that induce fixed mispredictions. In this work, we exploit an overlooked property of adversarial perturbations --- their dependence on image transforms --- and introduce transform-dependent adversarial attacks. Unlike traditional attacks, our perturbations exhibit metamorphic properties, enabling diverse adversarial effects as a function of transformation parameters.
We demonstrate that this transform-dependent vulnerability exists across different architectures (\eg, CNN and transformer), vision tasks (\eg, image classification and object detection), and a wide range of image transforms. Additionally, we show that transform-dependent perturbations can serve as a defense mechanism, preventing sensitive information disclosure when image enhancement transforms pose a risk of revealing private content.
Through analysis in blackbox and defended model settings, we show that transform-dependent perturbations achieve high targeted attack success rates, outperforming state-of-the-art transfer attacks by $17–31\%$ in blackbox scenarios. 
Our work introduces novel, controllable paradigm for adversarial attack deployment, revealing a previously overlooked vulnerability in deep networks.

\end{abstract}  
\section{Introduction}
\label{sec:intro}
Adversarial attacks on deep neural networks have traditionally been studied through the lens of  imperceptible perturbations that can deceive models into misclassifying inputs~\cite{goodfellow2014explaining,kurakin2016adversarial, madry2017towards,dong2018boosting,zou2020improving,wang2021admix,wang2023structure,xie2019improving,dong2019evading, lin2019nesterov}. In many real cases, inputs can undergo different transformations due to changes in viewpoint, lighting conditions, and resolution with little to no perceptual change. Prior work has attempted to make adversarial attacks robust to such input transformations by optimizing over the expectation of the attack objective under a distribution of transformations. For instance,  the expectation over transformation (EOT) framework \cite{athalye2018synthesizing} seeks transforma-invariant (or -independent) attacks that are robust to variation of input transformations.

\begin{figure}[t]
\centering
\small
\includegraphics[width=0.45\textwidth]{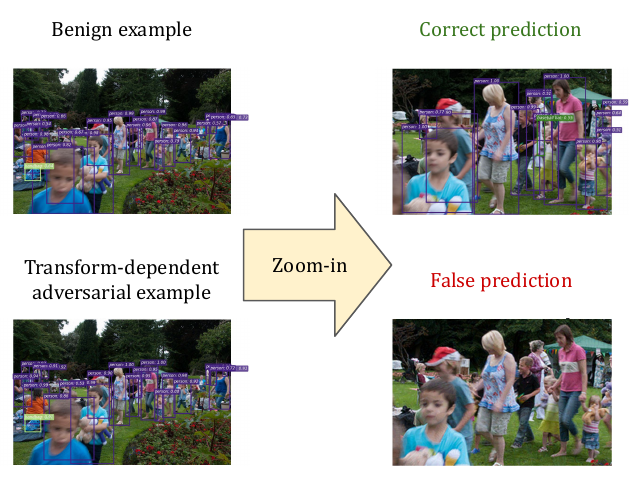}

\caption{This paper introduces transform-dependent adversarial attacks, where the adversarial effects are controllably triggered by image transforms, offering a flexibility for attack deployments or a protection against detection. 
In this example, our adversarial perturbation prevents persons from being detected by an object detector when zooming-in can potentially reveal the privacy details.}

\vspace{-5mm}

\label{fig:intro2}
\end{figure}
\begin{figure*}[t]
\centering
%
    \includegraphics[width=1.00\textwidth]{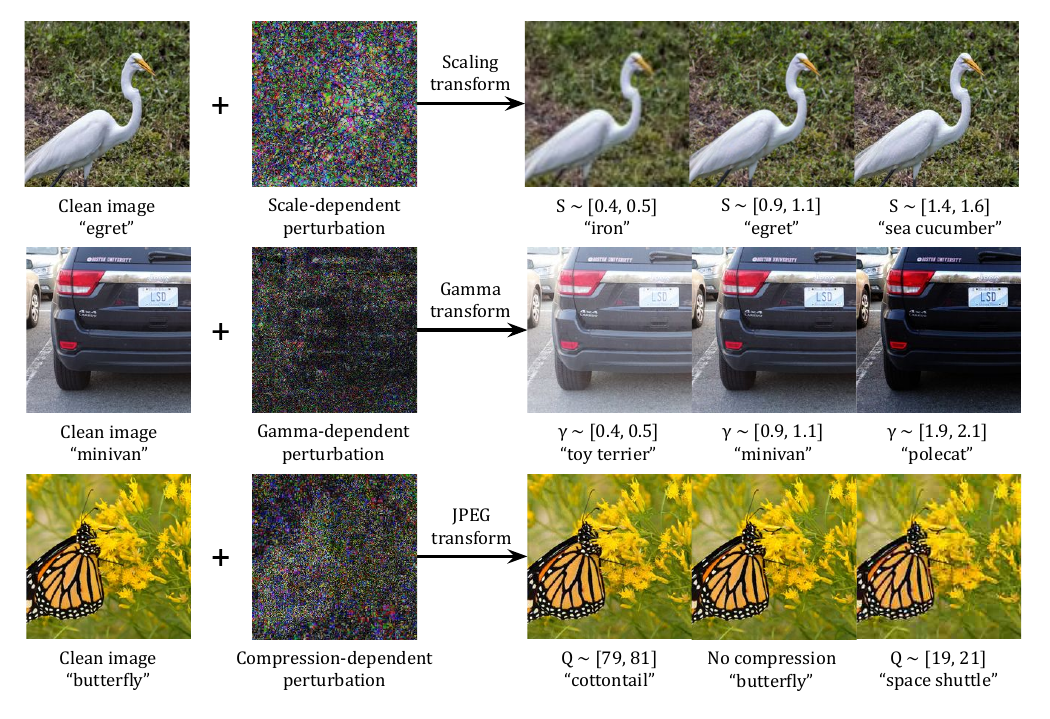}
\caption{Examples of transform-dependent adversarial attacks against classifiers. A single adversarial perturbation added to clean image can offer multiple attack effects for desired image transforms. 
\textit{First row}: Targeted attacks are triggered by scaling around $0.5\times$ and $2\times$, with clean label around $1\times$. Scaled images in the first row will have different sizes after scaling, but we present their resized versions for better display. 
\textit{Second row}: Attacks triggered with $\gamma \sim 0.5\pm 0.1, 2\pm 0.1$ in gamma correction,  while providing the clean label with $\gamma \sim 1\pm 0.1$. 
\textit{Third row}: Attacks triggered with JPEG image compression quality factor $Q \sim 80\pm1$, $20\pm1$, while providing the clean label with no compression.
\textit{The perturbation in all examples is bounded by $\ell_\infty \leq 8$; the magnitude is amplified $30\times$ for better visualization.} 
}
\label{fig:intro}
\end{figure*}

In this work, we explore an entirely new dimension of adversarial attacks --- transform-dependent adversarial attacks. Unlike conventional approaches that aim for invariance to transformations, we uncover a new threat: a single, carefully crafted perturbation can induce a wide range of targeted mispredictions, controlled by the applied transforms. This transform-dependent effect not only enhances stealthiness, extending beyond visually imperceptible noise, but also introduces a defense mechanism for images --- preventing sensitive information disclosure when image enhancement transforms risk exposing private content, as demonstrated in \cref{fig:intro2}.
%
%
%
In \cref{fig:intro}, we present examples of transform-dependent adversarial attacks, showing how a single perturbation can lead to different target labels depending on the transformation applied. Stealthiness is achieved by optimizing attack effects to remain latent under normal conditions, only being triggered when transformation parameters deviate from the original image state.

Our analysis of this new class of controlled, transform-dependent adversarial attacks reveals a profound threat for which current defenses are ill-prepared for. 
We demonstrate that a single additive perturbation can encode a remarkable diversity of targeted adversarial effects, each triggered by flexibly predefined input transforms. Unlike conventional attacks, which rely on isolated and static perturbations with fixed post-training effects, our approach introduces dynamic, controllable threats capable of manifesting in drastically different ways, depending on how the input is transformed. The expanded adversarial space introduced by image transforms highlights a significant vulnerability in modern deep networks from a new perspective.

\noindent We summarize our main contributions as follows.
\squishlist
\item We introduce the novel concept of transform-dependent adversarial attacks, where a single additive perturbation can embed multiple targeted mispredictions that are triggered by predefined transforms on the input image.
\item We showcase the versatility and practicality of proposed attacks through extensive experiments across models, tasks, and transforms, demonstrating how a single perturbation yields diverse, controllable adversarial effects.
\item Our study on defended and blackbox models suggests that transform-dependent attacks can achieve high blackbox transfer success rates and bypass common defenses. 
%
%
\item We demonstrate the real-world implications of transform-dependent perturbations as an image protection by extending our results to object detection, enabling selective hiding attacks based on image enhancement transforms.
\squishend

Our attack formulation represents the first of its kind effort to integrate multiple attack effects in a single adversarial perturbation, which can be dynamically activated in coordination with image transforms, enabling attacks deployment controllability.
Our work questions the fundamental assumption in adversarial machine learning --- that an input example is either benign or adversarial. Our work forces a conceptual paradigm shift, as adversarial perturbations can now encode metamorphic properties that reveal different attack effects based on transformations applied to the input. Defenses can no longer simply aim to detect static perturbations, but must grapple with the chameleon-like nature of our transform-dependent adversarial attacks. 
\section{Related Work}
\label{sec:related}

\subsection{Adversarial attacks and defenses}
Adversarial attacks were initially introduced as visually imperceptible perturbations, capable of causing false predictions in untargeted attacks or inducing specific misclassifications in targeted attacks \cite{goodfellow2014explaining, moosavi2017universal}. Adversarial vulnerabilities have been found across various neural architectures(\eg, CNN, transformer), and tasks (\eg, classification, object detection) \cite{xie2017adversarial,mahmood2021robustness}. In conventional paradigms, a single post-training perturbation can achieve only one fixed attack outcome, offering limited control over adversarial effects. Likewise, existing defenses operate under the assumption that an input is either benign or adversarial, without considering dynamic adversarial behaviors \cite{liao2018defense,raff2019barrage,dong2019evading}.
In this work, we challenge this paradigm by leveraging image transforms in adversarial optimization. Our proposed transform-dependent attacks introduce a new level of controllability, allowing a single perturbation to induce multiple, transform-driven attack effects. This approach not only expands the adversarial landscape but also effectively bypasses common defenses.

\subsection{Image transformation in adversarial attacks}\label{subsec:related-trans-attack}
In adversarial attack literature, image transforms have primarily been leveraged for two purposes: 1) Generating transform-invariant adversarial examples that remain effective under input corruptions \cite{athalye2018synthesizing,kurakin2018adversarial,Lennon_2021_ICCV}, and transferable adversarial examples that remain adversarial across models that are differ from the model used in the attacks generation process \cite{zou2020improving,wang2021admix,wang2023structure,xie2019improving,dong2019evading,lin2019nesterov}. This line of work seeks robust adversarial examples that provide consistent attack effects under various conditions.
2) Generating adversarial examples via simple geometric transforms~\cite{pei2017towards, engstrom2019exploring, xiao2018spatially, chen2020explore}. Without optimizing a $\ell_\infty$-norm bounded perturbation, this line of work optimizes transformation parameter, so that a slightly transformed image causes misprediction. While they provide new attacking form against deep networks through simple image transforms, the attacks are limited to certain transformation parameters and some requires to change the image semantic content~\cite{xiao2018spatially} or deviated from natural image color distribution~\cite{chen2020explore}, which limits their practicality.
In contrast, our work exploits the more general and diverse image transformations, covering spatial, photometric, and compression transformations to craft adversarial perturbations capable of dynamically altering their effects based on the transform applied. Introducing controllable attack effects by transforming the input in various ways.


\section{Method}
\label{sec:method}

\subsection{Preliminaries} \label{subsec:method-prelim}
We consider additive perturbations for adversarial attacks~\cite{szegedy2013intriguing,goodfellow2014explaining,madry2017towards} that generate an adversarial example as $\mathbf{x}+\delta$, where $\delta$ is an adversarial perturbation for a given image $\mathbf{x}$. Given a well-trained model $f$ that provides correct prediction, $f(\mathbf{x})=y$. We can learn $\delta$ for an untargeted adversarial attack  such that $f(\mathbf{x}+\delta) \neq y$ (i.e., the  prediction does not match the correct label) or a targeted attack such that $f(\mathbf{x}+\delta) = y^\star$  (i.e., the prediction is the desired target label $y^\star$). To keep the perturbation imperceptible, $\delta$ is usually bounded within the $\ell_p$ norm ball $\|\delta\|_p \leq \varepsilon$. In general, targeted adversarial attacks can be generated by solving the following optimization problem:
\begin{equation}\label{eq:attack}
    \min_{\delta}~\mathcal{L}(f(\mathbf{x}+\delta), y^\star)\quad \text{s.t. } \|\delta\|_p \leq \varepsilon.
\end{equation}

A general choice for the loss function $\mathcal{L}$ is the training loss for the corresponding tasks. Several algorithms have been proposed to solve this optimization problem; notable examples include  FGSM~\cite{goodfellow2014explaining}, PGD~\cite{madry2017towards}, MIM~\cite{dong2018boosting} and Auto-Attack~\cite{croce2020reliable, croce2021mind}. In this work, we focus on PGD attacks for their simplicity and effectiveness. PGD iteratively solves the attack optimization problem as 
\begin{equation}\label{eq:PGD}
    \delta^{t+1} = \Pi_{\varepsilon} \left( \delta^{t} - \alpha~ \mathbf{sign} (\nabla_{\delta}\mathcal{L}(f(\mathbf{x}+\delta^{t}), y^\star)) \right),
\end{equation}
where $\delta^t$ denotes perturbation at iteration $t$ that is updated using sign of gradient with step size $\alpha$. $\Pi_{\varepsilon}$ denotes an operator that projects the updated $\delta^t$ back to the $\ell_p$-norm ball and obtain $\delta^{t+1}$.

\subsection{Transform-dependent attacks}
Let us define the image transform function as $T(\mathbf{x};\theta)$ that transforms the input image $\mathbf{x}$  according to the given transform parameter $\theta \in \Theta$. Applying the transform function over clean input images does not cause significant degradation in the accuracy of networks that are properly trained with data augmentation techniques~\cite{rowley1998rotation,jaderberg2015spatial, marcos2017rotation}. In other words, the output of the transformed images remains same as the original image:   
\begin{equation}\label{eq:transform-benign}
    f(T(\mathbf{x}; \theta)) \sim y.
\end{equation}

To introduce image transform-dependent effects in adversarial examples, we incorporate the transform function in the attack generation. Specifically, we aim for transform-dependent targeted attacks with target label $y^\star_i$ for the corresponding transform parameter $\theta_i$.

In the most general form, we assume $\boldsymbol  \theta$ and $\mathbf{y}^\star$ represent two vectors with $N$ discrete targets embedded in the transform-dependent attacks as $\boldsymbol\theta = \{\theta_i\}_{i=1}^N$, $\mathbf{y}^\star = \{y^\star_i\}_{i=1}^N$. 
The attacker can select a mapping for \textbf{parameter-target pairs} that trigger the label $y_i^\star$ for a given transform parameter $\theta_i$, or choose the $y_i^\star$ as true label $y_i$ for certain transform parameters where the adversarial examples are intended to act as ``benign'' for better stealthiness. 
%
%
We seek to generate a single transform-dependent perturbation $\delta$ as a solution of the following optimization problem:
\begin{equation}\label{eq:attack-transform-dep}
    \min_\delta~\sum_i \mathcal{L}(f(T(\mathbf{x}+\delta; \theta_i)), y^\star_{i}) \quad {\text{s.t. }\|\delta\|_p \leq \varepsilon}. 
\end{equation}
At the test time, the perturbed image can be created as $\mathbf{x} + \delta$, the transform-dependent adversarial examples can be created as $T(\mathbf{x} + \delta;\theta_i)$, and attacks can be triggered by providing malicious $\theta_i$, resulting in transform-dependent effects.

To enhance the robustness of transform-dependent attacks against small variations in transform parameters, we adopt the approach from EOT \cite{athalye2018synthesizing}, optimizing perturbations that remain effective under slight parameter changes.
We achieve this by incorporating expectation over transforms (EOT)-based data augmentation into the optimization process. Specifically, we seek a perturbation $\delta$ that consistently induces the target label $y^\star_i$ across a range of transform parameters $\theta_i$ within a neighborhood of $\bar \theta_i$. This is formulated as:
\begin{equation}\label{eq:attack-transform-dep-range}
    \min_\delta~\underset{\theta_i \sim N_r(\bar \theta_i)}{\mathbb{E}}~\sum_i \mathcal{L}(f(T(\mathbf{x}+\delta; \theta_i)), y^\star_{i}) \quad {\text{s.t. }\|\delta\|_p \leq \varepsilon}, 
\end{equation}
where $N_r(\bar \theta) = \{ \theta \in \Theta \mid  \left\| \theta - \bar \theta \right\| < r \}$ represents a uniform distribution around $\bar \theta$ with radius $r$. This ensures the perturbation remains effective across a continuous range of parameter values rather than discrete points.

We also observe that perturbations optimized over transform parameter ranges for multiple targets successfully transfer to blackbox models. This aligns with prior findings \cite{lin2019nesterov,xie2019improving,wang2021admix}, which highlight the role of image transformations in enhancing perturbation transferability. As detailed in \cref{sec:transfer}, transform-dependent perturbations crafted on a single surrogate model can effectively attack multiple blackbox models with different architectures. Moreover, these attacks preserve their transform-dependent properties, deceiving blackbox models into predicting the intended target labels based on the applied transformation.

\subsection{Transform functions}\label{subsec:method-transform}
We focus on transform functions that are both differentiable and deterministic. Differentiability ensures compatibility with gradient-based attack algorithms, allowing the loss gradient with respect to $\delta$ to be expressed as
$
    \nabla_{\delta}\mathcal{L}(f(T(\mathbf{x}+\delta; \theta)), y^\star) = 
    \frac{\partial \mathcal{L}}{\partial f}
    \frac{\partial f}{\partial T}
    \frac{\partial T}{\partial \delta}. 
$
Deterministic transforms ensure that modifications are precisely controlled by $\theta$ rather than applied randomly, allowing attackers to manipulate transform-dependent effects with precision.

Based on these two properties, we adopt a variety of commonly used image transforms, covering spatial, photometric, and compression-based modifications. Our experiments primarily use scaling, blurring, and gamma correction, but the approach generalizes to other differentiable transforms. Scaling, parameterized by a factor $S$, resizes an image $\mathbf{x} \in \mathbb{R}^{H\times W\times3}$ into $T(\mathbf{x}; S) \in \mathbb{R}^{S H\times S W\times3}$ using differentiable interpolations such as bilinear or bicubic methods. Gaussian blurring $T(\mathbf{x}; \sigma)$ is controlled by the standard deviation $\sigma$ of the blur kernel, while gamma correction applies brightness adjustment with $T(\mathbf{x}; \gamma) = A \mathbf{x}^{\gamma}$, where $A$ is a constant that normalizes intensity. Though JPEG compression is non-differentiable due to quantization, we use the differentiable approximation $JPEG_{\text{diff}}$~\cite{shin2017jpeg} to enable transform-dependent attacks, with compression $Q$ as transform parameter, which is formulated as $T(\textbf{x};Q) = JPEG_{\text{diff}}(\textbf{x},Q)$.

These transforms serve as effective adversarial mechanisms, allowing targeted perturbations to exploit model vulnerabilities in a transformation-aware manner.


\section{Experiments}
\label{sec:experiment}
In this section, we first demonstrate the effectiveness of transform-dependent attacks across a wide range of transforms, and image classifier architectures. Later, we analyze our attacks in blackbox and defended model scenarios, showcasing its competitive performance to state-of-the-art methods that are specifically designed for blackbox transferability. Additionally, we highlight the generalizability of our attack formulation to object detection, emphasizing its practical applicability and potential use cases.

\begin{table*}[t]
\centering
\small
\caption{
Transform-dependent targeted ASR (\%) $\uparrow$ against classifiers over the range of selected parameters. A higher ASR value indicates better attack performance. The adversarial perturbation budget for this experiment is  $\varepsilon=8$.
}
\label{tab:cls-asr-range}

\begin{tabular}{ccccccccc}
\hline
\multicolumn{1}{c|}{\multirow{2}{*}{\begin{tabular}[c]{@{}c@{}}Transform\\ parameter\end{tabular}}} & \multicolumn{8}{c}{Classifier model} \\
\multicolumn{1}{c|}{} & VGG19 & ResNet50 & Dense121 & Incv3 & Mobv2 & ViT-L16 & ViT-L32 & Swin-T \\ \hline
\multicolumn{1}{c|}{$S \sim [0.4,0.6]$} & 95.80 & 87.20 & 88.20 & 61.90 & 96.40 & 68.70 & 61.00 & 99.70 \\
\multicolumn{1}{c|}{$S \sim [0.9,1.1]$} & 99.90 & 98.70 & 98.20 & 83.10 & 100.0 & 75.50 & 68.40 & 100.0 \\
\multicolumn{1}{c|}{$S \sim [1.4,1.6]$} & 99.90 & 99.60 & 99.60 & 80.30 & 99.80 & 75.90 & 68.70 & 100.0 \\ \hline
\multicolumn{1}{c|}{Average} & 98.53 & 95.17 & 95.33 & 75.10 & 98.73 & 73.37 & 66.03 & 99.90 \\ \hline
\multicolumn{1}{c|}{$\sigma \sim [0.4,0.6]$} & 100.0 & 99.90 & 99.80 & 95.40 & 99.80 & 97.40 & 91.70 & 100.0 \\
\multicolumn{1}{c|}{$\sigma \sim [1.4,1.6]$} & 98.70 & 98.90 & 97.90 & 75.40 & 95.80 & 77.60 & 66.20 & 98.90 \\
\multicolumn{1}{c|}{$\sigma \sim [2.9,3.1]$} & 97.10 & 98.90 & 94.40 & 65.70 & 94.20 & 65.60 & 53.80 & 99.10 \\ \hline
\multicolumn{1}{c|}{Average} & 98.60 & 99.23 & 97.37 & 78.83 & 96.60 & 80.20 & 70.57 & 99.33 \\ \hline
\multicolumn{1}{c|}{$\gamma \sim [0.4,0.6]$} & 99.90 & 100.0 & 99.90 & 93.10 & 99.90 & 99.00 & 93.30 & 99.80 \\
\multicolumn{1}{c|}{$\gamma \sim [0.9,1.1]$} & 100.0 & 96.20 & 99.90 & 95.10 & 99.90 & 97.60 & 91.90 & 100.0 \\
\multicolumn{1}{c|}{$\gamma \sim [1.9,2.1]$} & 100.0 & 91.00 & 99.70 & 93.70 & 99.70 & 93.60 & 85.30 & 99.70 \\ \hline
\multicolumn{1}{c|}{Average} & 99.97 & 95.73 & 99.83 & 93.97 & 99.83 & 96.73 & 90.17 & 99.83 \\ \hline
\multicolumn{1}{c|}{$Q \sim [19,21]$} & 96.10 & 87.70 & 91.70 & 73.40 & 86.90 & 66.60 & 72.70 & 83.40 \\
\multicolumn{1}{c|}{$Q \sim [49,51]$} & 99.10 & 96.20 & 96.70 & 80.90 & 95.20 & 81.90 & 83.00 & 96.20 \\
\multicolumn{1}{c|}{$Q \sim [79,81]$} & 99.40 & 98.40 & 99.20 & 87.00 & 97.90 & 87.80 & 82.20 & 99.80 \\ \hline
\multicolumn{1}{c|}{Average} & 98.20 & 94.10 & 95.87 & 80.43 & 93.33 & 78.77 & 79.30 & 93.13 \\ \hline
\end{tabular}

\end{table*}
\begin{figure*}[!h]
\centering
    \includegraphics[width=1.0\linewidth]{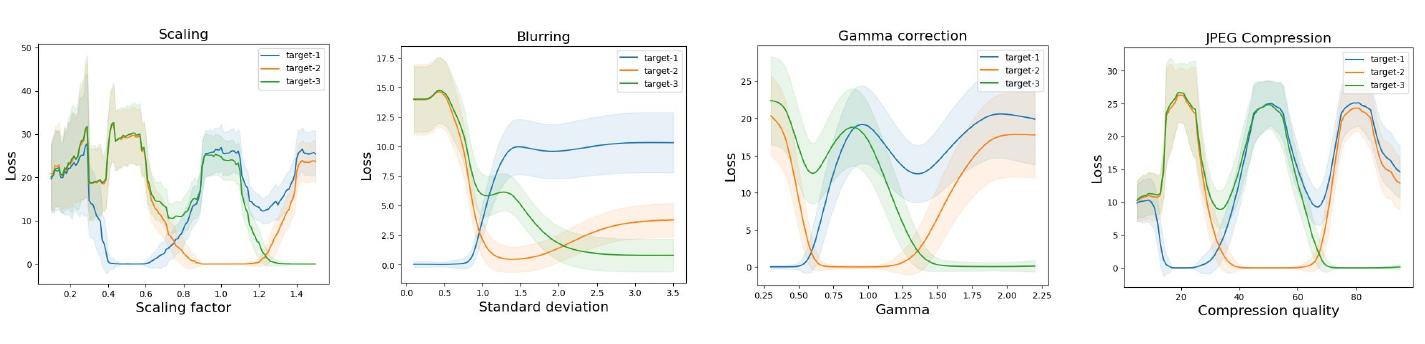}
\caption{Loss landscape of the ResNet50 whitebox model over transform parameter values. A small loss value indicates successful targeted attacks within the desired transform parameter ranges. 
Perturbation are generated to deceive model providing three target labels for three transform parameter ranges: $S\sim\{[0.4,0.6], [0.9,1.1], [1.4,1.6]\}$, $\sigma\sim\{[0.4,0.6], [1.4,1.6], [2.9,3.1]\}$, $\gamma\sim\{[0.4,0.6], [0.9,1.1], [1.9,2.1]\}$, $Q\sim\{[19,21], [49,51], [79,81]\}$, consistent setup as \cref{tab:cls-asr-range}. This figure suggests that multiple attack targets can be controllably triggered by transform parameters, and attacks remain effective when parameter is sampled outside of ranges.
}

\label{fig:loss}
\end{figure*}

\begin{table*}[ht]
\centering
\caption{Blackbox transfer evaluation under untargeted and targeted settings, adversarial perturbation budget is $\varepsilon=8$. Our scale-dependent attacks achieve comparable untargeted ASRs and higher targeted ASRs under the same setting as the most recent transfer attacks. Furthermore, the transfer attacks maintain transform-dependent attack properties as targeted ASRs present.
}
\label{tab:attack-transfer-bb}
\small
\begin{tabular}{c|ccccc|ccccc}
\hline
\multirow{3}{*}{Methods} & \multicolumn{5}{c|}{Untargeted ASR (\%) $\uparrow$} & \multicolumn{5}{c}{Targeted ASR (\%) $\uparrow$} \\ \cline{2-11} 
 & \multicolumn{1}{c|}{Surrogate} & \multicolumn{4}{c|}{Blackbox model} & \multicolumn{1}{c|}{Surrogate} & \multicolumn{4}{c}{Blackbox model} \\
 & \multicolumn{1}{c|}{ResNet50} & VGG19 & Dense121 & Incv3 & Mobv2 & \multicolumn{1}{c|}{ResNet50} & VGG19 & Dense121 & Incv3 & Mobv2 \\ \hline
BPA~\cite{xiaosen2023rethinking} & \multicolumn{1}{c|}{99.40} & 60.96 & 70.70 & 35.36 & 68.90 & \multicolumn{1}{c|}{100.0} & 31.02 & 43.82 & 15.34 & 39.00 \\ \hline
ILPD~\cite{li2024improving} & \multicolumn{1}{c|}{83.96} & \textbf{88.10} & \textbf{90.68} & 64.70 & - & \multicolumn{1}{c|}{-} & - & - & - & - \\ \hline
Logit-SU~\cite{wei2023enhancing} & \multicolumn{1}{c|}{-} & - & - & - & - & \multicolumn{1}{c|}{-} & 41.30 & 45.70 & 1.10 & - \\ \hline
$S \sim [0.4,0.6]$ & \multicolumn{1}{c|}{88.00} & 80.10 & 81.10 & \textbf{96.90} & \textbf{86.80} & \multicolumn{1}{c|}{99.60} & 39.10 & 40.00 & 12.10 & 31.50 \\
$S \sim [0.9,1.1]$ & \multicolumn{1}{c|}{98.30} & 86.70 & 77.50 & 62.60 & 81.00 & \multicolumn{1}{c|}{99.90} & \textbf{62.80} & 61.40 & 28.80 & 56.10 \\
$S \sim [1.4,1.6]$ & \multicolumn{1}{c|}{99.90} & 85.40 & 83.90 & 58.90 & 85.10 & \multicolumn{1}{c|}{100.0} & 59.50 & \textbf{67.80} & \textbf{33.10} & \textbf{60.00} \\ \hline
$\sigma \sim [0.4,0.6]$ & \multicolumn{1}{c|}{99.80} & 64.40 & 69.40 & 53.50 & 65.60 & \multicolumn{1}{c|}{100.0} & 32.60 & 42.20 & 19.10 & 35.00 \\
$\sigma \sim [1.4,1.6]$ & \multicolumn{1}{c|}{99.80} & 85.40 & 84.60 & 73.70 & 83.80 & \multicolumn{1}{c|}{99.90} & 43.90 & 52.60 & 23.40 & 35.40 \\
$\sigma \sim [2.9,3.1]$ & \multicolumn{1}{c|}{99.60} & 86.80 & 85.50 & 76.30 & 85.00 & \multicolumn{1}{c|}{99.90} & 41.90 & 50.40 & 22.50 & 33.80 \\ \hline
\end{tabular}

\end{table*}

\subsection{Experiment setup}
\noindent\textbf{Models and dataset.} 
We utilize pretrained image classification models from Pytorch Torchvision \cite{paszke2017automatic}, which provides models pretrained on ImageNet dataset~\cite{deng2009imagenet} for a variety of families and architectures. Models are trained with data augmentation techniques including random cropping, rotation, flipping, and color jittering, so that they align with our assumption in \cref{eq:transform-benign}. We sample models from different families that cover Convolutional Neural Networks (CNNs) and Vision Transformer (ViT): \{VGG-19-BN~\cite{simonyan2014very}, ResNet-50~\cite{he2016deep}, DenseNet-121~\cite{huang2017densely}, InceptionV3~\cite{szegedy2016rethinking}, MobileNet-v2~\cite{sandler2018mobilenetv2}, ViT-L-16, ViT-L-32~\cite{dosovitskiy2020image}, Swin-T~\cite{liu2021swin}\}. We use $1000$ ImageNet-like RGB images from the NeurIPS17' challenge~\cite{neurips17challenge}, which has $1000$ classes same as ImageNet and all images in the size of $224\times224$.

\noindent\textbf{Attack settings.} We focus on targeted attacks, as they are more challenging than untargeted ones and align with our transform-dependent attack formulation. For simplicity, we present results for $N=3$ transform parameter-target pairs: $\{\bar{\theta_i}\}_{i=1}^3$ and $\{y^\star_i\}_{i=1}^3$, in the main text. We provide further analysis for different choices of $N$ in \cref{sec:additional}. The goal is to deceive the model into providing three distinct, desired target labels that are randomly selected from ImageNet classes, when the victim model is provided with the corresponding transformed perturbed images. For optimization, we used $500$ iterations for PGD with the step size of $\alpha=5\times 10^{-4}$ for sufficient convergence.

\noindent\textbf{Evaluation metrics.} 
We evaluate the transform-dependent attack performance using the Attack Success Rate (ASR). Specifically, for a perturbed image transformed as $T(\mathbf{x}+\delta;\theta_i)$, we check whether the network prediction matches target label $y^*_i$. We report ASR for each transform parameter $\bar{\theta_i}$ over $N_r(\bar \theta_i)$, where $N_r(\bar \theta_i)$ includes series of $\theta$ samples generated with a small sampling rate (detailed in \cref{subsec:whitebox}), and we report ASR averages over all samples within $N_r(\bar \theta_i)$ for each $\bar{\theta_i}$.

\subsection{Transform-dependent attacks on classifiers}
\label{subsec:whitebox}
We present transform-dependent attacks using four image transformations commonly applied in real-world scenarios (as described in \cref{subsec:method-transform}): geometric transforms (\textbf{scaling}, \textbf{blurring}) and photometric transforms (\textbf{gamma correction}, \textbf{JPEG compression}).
For each transform, we selected three parameters $\{\bar\theta_i \}$ based on the criterion of minimizing the impact on model accuracy in the absence of perturbation.

For scaling, we use factors $S \in [0.5, 1.0, 1.5]$. For blurring, we fix the Gaussian kernel size to $5\times5$ and vary $\sigma \in [0.5, 1.5, 3.0]$. For gamma correction, we select $\gamma \in [0.5, 1.0, 2.0]$, and for JPEG compression, we use quality levels $Q \in [20,50,80]$. To generate $N_r(\bar \theta_i)$, we define a neighborhood with an interval radius of $r=0.1$ for scaling, blurring, gamma correction, and $r=1$ for JPEG compression. This results in the following parameter ranges: scaling $S \sim \{[0.4,0.6], [0.9,1.1], [1.4,1.6]\}$, blurring $\sigma \sim \{[0.4,0.6], [1.4,1.6], [2.9,3.1]\}$, gamma correction $\gamma \sim \{[0.4,0.6], [0.9,1.1], [1.9,2.1]\}$, and JPEG compression $Q \sim \{[19,21], [49,51], [79,81]\}$. 

\cref{tab:cls-asr-range} summarizes the ASR for transform-dependent attacks across three target labels and corresponding parameter ranges. These results show consistent targeted attacks success on whitebox CNN and ViT models, demonstrating the effectiveness of transform-dependent adversarial examples. These attacks, optimized via \cref{eq:attack-transform-dep-range}, retain their effectiveness under small parameter variations (also indicated in \cref{fig:loss}). Their transferability to blackbox models is further explored in \cref{sec:transfer}.
\cref{fig:loss} visualizes the adversarial loss landscape of ResNet50 across four transform-dependent attacks, with loss evaluated over a grid of transform parameters (sampling rate $0.1$ for $S,\sigma,\gamma$, and $1$ for $Q$). The three colors represent distinct target labels, with solid lines indicating average loss and shaded areas denoting standard deviation. As seen, minimum loss values align with the intended transform parameter ranges, confirming that adversarial examples successfully embed targeted attacks triggered by specific transformations. Notably, scaling and JPEG are more sensitive to parameter variations, while blurring and gamma correction exhibit greater smoothness.



\begin{table}[t]


\centering
\caption{Results of untargeted ASR (\%) $\uparrow$ against commonly used defense methods, perturbation budget is $\varepsilon=8$. 
}
\label{tab:attack-defense}
\footnotesize

\begin{tabular}{c|cccc}
\hline
\multirow{2}{*}{Attack method} & \multicolumn{4}{c}{Defense method} \\
 & HGD~\cite{liao2018defense} & RS~\cite{cohen2019certified} & JPEG~\cite{guo2017countering} & NPR~\cite{naseer2020self} \\ \hline
BPA~\cite{xiaosen2023rethinking} & 23.96 & 14.00 & 22.52 & 14.08 \\ \hline
Scaling (ours) & 56.20 & 53.43 & 34.90 & 39.70 \\
Blurring(ours) & 57.73 & 63.67 & 57.80 & 52.43 \\
Gamma(ours) & 48.67 & 65.43 & 52.27 & 53.57 \\ \hline

\end{tabular}

\vspace{-4.3mm}

\end{table}
\begin{figure*}[ht]
\centering
%
    \includegraphics[width=1.0\textwidth]{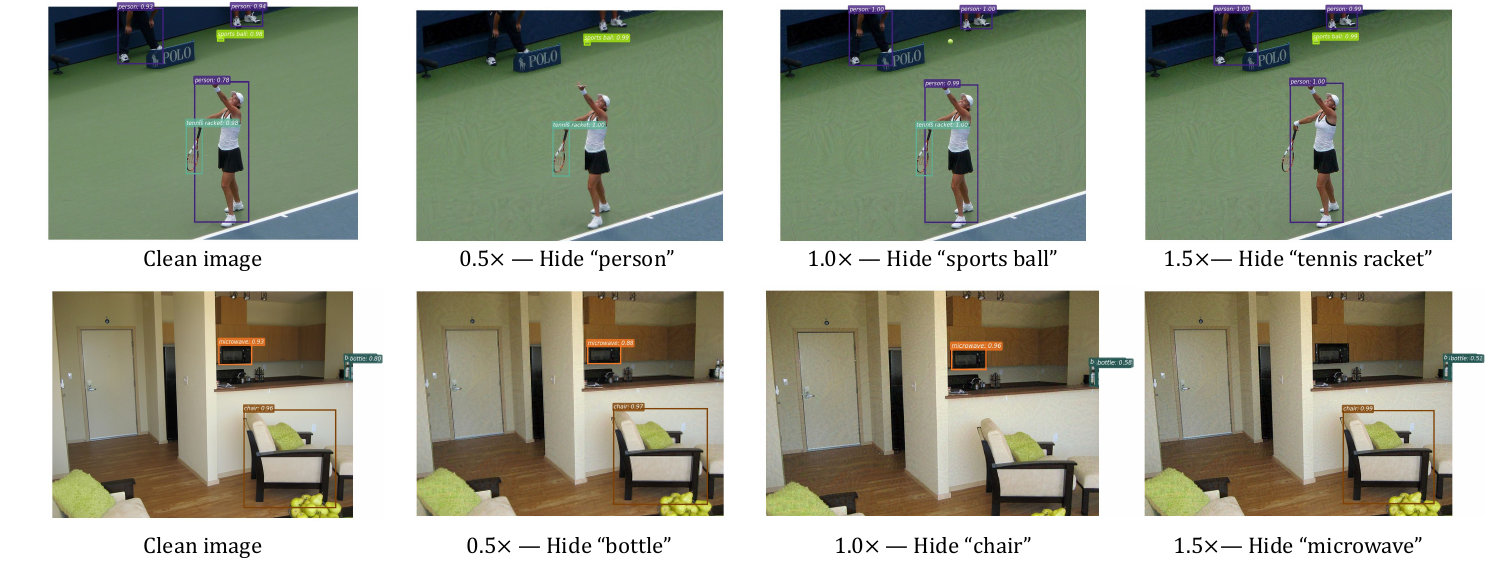}

\vspace{-3mm}
\caption{Visualization of scale-dependent selective hiding attack against YOLOv3.
The first column shows detection results on the original clean image, while the following columns present perturbed images scaled with factors $S\in\{0.5,1.0,1.5\}$ (perturbation $\|\delta\|_\infty \leq 10$). \textit{Note that scaled images will have different sizes after scaling, but we present their resized versions for better display}.
}
\label{fig:det}
\end{figure*}
\begin{table*}[h]

\centering
\small
\caption{Quantitative evaluation of scale-dependent selective hiding attack success rate (ASR) over object detection models. The higher value indicates better attack performance. Our attack formulation is generalizable to more complex object detection task.
}
\label{tab:attack-detector}

\begin{tabular}{c|ccccc|ccccc}
\hline
\multirow{2}{*}{\begin{tabular}[c]{@{}c@{}}Scaling \\ factor\end{tabular}} & \multicolumn{5}{c|}{\begin{tabular}[c]{@{}c@{}}Detector model ASR (\%) $\uparrow$\\ $\|\delta\|_\infty \leq 10$\end{tabular}} & \multicolumn{5}{c}{\begin{tabular}[c]{@{}c@{}}Detector model ASR (\%) $\uparrow$\\ $\|\delta\|_\infty \leq 20$\end{tabular}} \\
 & Faster & YOLOv3 & FCOS & GRID & DETR & Faster & YOLOv3 & FCOS & GRID & DETR \\ \hline
$S=0.5$ & 53.85 & 97.39 & 71.61 & 54.44 & 11.91 & 60.26 & 97.39 & 76.39 & 67.22 & 17.16 \\
$S=1.0$ & 65.71 & 95.10 & 69.38 & 58.89 & 11.95 & 72.76 & 98.04 & 79.26 & 68.52 & 23.61 \\
$S=1.5$ & 90.71 & 32.35 & 100.0 & 86.11 & 41.26 & 96.15 & 40.98 & 100.0 & 97.78 & 52.18 \\ \hline
Average & 70.09 & 74.95 & 80.33 & 66.48 & 21.71 & 76.39 & 78.80 & 85.22 & 77.84 & 30.98 \\ \hline
\end{tabular}

\end{table*}

\begin{figure*}[ht]
\centering
%
    \includegraphics[width=1.0\textwidth]{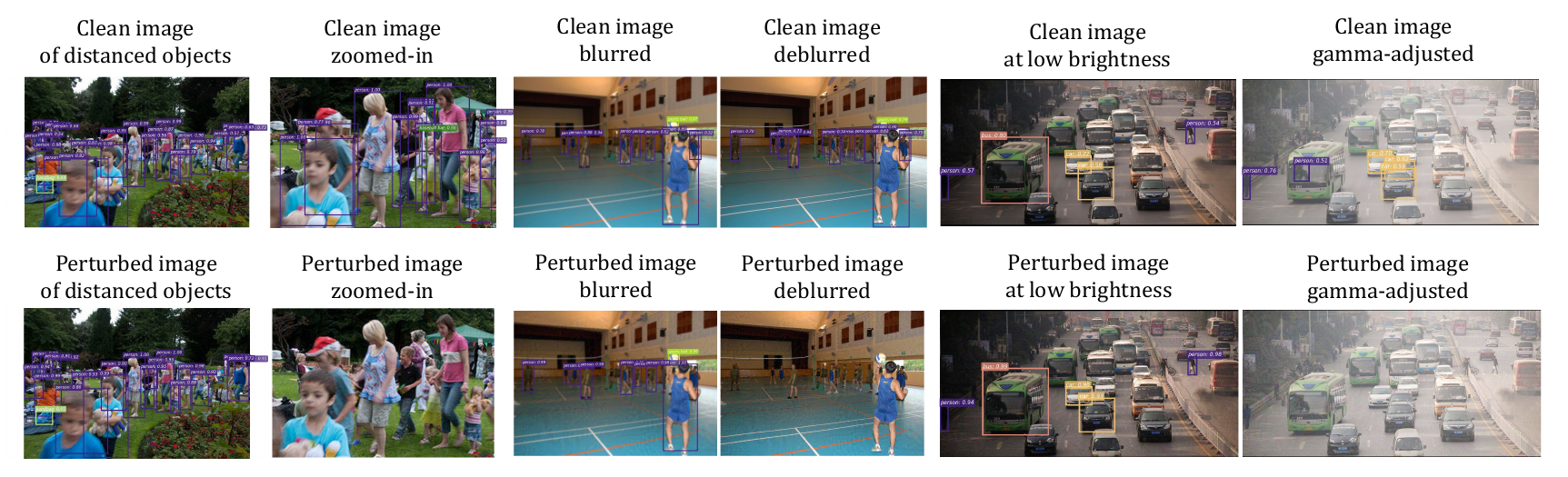}
\caption{Visualization of the enhance transform-hiding attack on YOLOv3. While objects in the enhanced clean images are being detected, after adding enhance transform-dependent perturbations ($\|\delta\|_\infty \leq 10$), detector fails when the enhancement transform is applied, preventing sensitive information disclosure in remote sensing or public surveillance systems. 
}
\label{fig:det-enhance}
\end{figure*}
\begin{table*}[h]

\centering
\small
\caption{Transform-selective hiding attack success rate (ASR) and accuracy (ACC) on object detectors. Higher ASR and ACC indicate better attack performance. The attack is selectively triggered within the desired range of enhancement transform parameters. Our attack consistently hides objects under image enhancement transforms while preserving detectability in unaltered or minimally enhanced images.
}
\label{tab:attack-detector-enhance}

\begin{tabular}{c|cccc|cccc|cccc}
\hline
\multirow{2}{*}{\begin{tabular}[c]{@{}c@{}}Transform\\ range\end{tabular}} & \multicolumn{4}{c|}{Zoom-in} & \multicolumn{4}{c|}{Deblurring} & \multicolumn{4}{c}{Gamma correction} \\
 & Faster & YOLOv3 & GRID & DETR & Faster & YOLOv3 & GRID & DETR & Faster & YOLOv3 & GRID & DETR \\ \hline
Attack & 89.41 & 95.22 & 84.62 & 53.79 & 78.92 & 85.17 & 72.37 & 51.03 & 79.88 & 89.10 & 72.13 & 53.46 \\ \hline
Safe & 94.36 & 99.03 & 93.25 & 77.02 & 92.43 & 91.78 & 88.84 & 71.53 & 89.49 & 91.78 & 82.50 & 65.96 \\ \hline
\end{tabular}

\end{table*}

\subsection{Attacks against blackbox and defended models} \label{sec:transfer}

\noindent\textbf{Blackbox transferability.}
Transform-dependent attacks leverage adversarial perturbations as a function of transform parameters, enabling effective blackbox attacks. Since only model outputs are accessible in the blackbox setup, we query blackbox models with transformed versions of perturbed images $T(x+\delta; \theta)$ over $N_r(\bar \theta_i)$, with three queries per image at maximum, which is a negligible query cost.
For an adversarial example $x+\delta$ generated for a surrogate model, we sample transform parameters $\theta$ from the neighborhoods used in \cref{tab:cls-asr-range}, \ie, $N_r(\bar \theta_i)$. The goal is to find a transformation $\theta_i^\star$ such that the adversarial example $T(x+\delta; \theta_i^\star)$ deceives the blackbox model:

\begin{equation}\label{eq:attack-transform-dep-bb}
    \text{Find } \theta_i^* ~~~ \text{s.t.}~~~f_{\text{bb}}(T(x+\delta;\theta_i^\star)) = y_i^\star;~  \theta_i^\star \in N_r(\bar \theta_i).
\end{equation}

We conduct a uniform search over $N_r(\bar \theta_i)$ with a sampling interval of $0.1$, leading to three queries per transform parameter (\eg, $\{0.4,0.5,0.6\}$ for $S=0.5$). An attack is successful if a transformed adversarial example forces the blackbox model to predict the target label.

\cref{tab:attack-transfer-bb} presents the transfer attack success rates, comparing our method against most-recent transfer attacks BPA~\cite{xiaosen2023rethinking}, SU~\cite{wei2023enhancing}, and ILPD~\cite{li2024improving}. We evaluate attacks under both targeted and untargeted setups. By introduce multi-targets in the adversarial optimization, our method achieves comparable ASR to recent techniques without specialized adaptation.
Scale- and blur-dependent attacks achieve high success in both targeted and untargeted setups. For untargeted attacks, we use the ground-truth label ($y$) as the target in \cref{eq:attack-transform-dep} for all transform parameters $\theta_i$ and maximize the adversarial loss. Scale-dependent attacks exhibit slightly better transferability than blur-dependent ones. Targeted success is higher for larger scaling factors, while untargeted success occurs more frequently at smaller scaling factors. These findings highlight that leveraging image transforms and embedding multi-targets improves blackbox transferability, demonstrating the effectiveness of our attacks.

\noindent\textbf{Attacks against defended models.}
To further evaluate the effectiveness of the transform-dependent attacks, we assess their performance on four defense methods:
HGD~\cite{liao2018defense}, Randomized Smoothing (RS)~\cite{cohen2019certified}, JPEG compression (JPEG)~\cite{guo2017countering} and NPR~\cite{naseer2020self}.
We follow the untargeted setup in a recent transfer attacks BPA~\cite{xiaosen2023rethinking}, generate whitebox attacks with perturbation budget $\varepsilon=8$, against four defenses applied to the ResNet50 model.
In \cref{tab:attack-defense}, we report these untargeted ASRs averaged over all the transform parameters. Our attack archives overall better performance than BPA on the same benchmark. These results suggest that transform-dependent attacks can bypass existing defense methods by leveraging the transformation space.

\subsection{Extension to object detection task}

To demonstrate the practicality of transform-dependent attacks, we design and present two attack scenarios on object detectors: object-selective and transform-selective hiding attacks. We first outline the experimental setup, followed by a discussion of the threat model and results.  

\noindent \textbf{Models and dataset.} 
For object detection, we use pretrained models from \texttt{MMDetection}~\cite{mmdetection} on the COCO 2017 dataset~\cite{lin2014microsoft}, and generate adversarial examples using the same dataset. Our model selection covers diverse architectures: one-stage detectors (YOLOv3~\cite{redmon2018yolov3}, FCOS~\cite{tian2019fcos}), two-stage detectors (Faster R-CNN~\cite{ren2015faster}, Grid R-CNN~\cite{lu2019grid}), and a ViT-based model (DETR~\cite{carion2020end}).

\noindent \textbf{Object-selective hiding.} We design a scenario where specific object classes are hidden when images are displayed at different scales, simulating real-world applications such as privacy-preserving surveillance or content-adaptive filtering. For example, certain sensitive objects (\eg, license plates or faces) could be concealed at lower resolutions in public monitoring systems, while critical details remain visible at higher resolutions for authorized analysis. In this setup, we use images containing three distinct classes and aim to hide objects from one of these classes for each transform parameter.
We focus on scaling with factors $S\in\{0.5, 1.0, 1.5\}$. The attack conceals objects from class $A$ at $0.5\times$, class $B$ at $1.0\times$, and class $C$ at $1.5\times$. ASR is measured as the ratio of successfully hidden objects in the final detection results for each scaled input.

\cref{tab:attack-detector} show that on most of R-CNN-based detectors, the scale-dependent attacks are successfully triggered when adversarial examples scale to the predefined image size, with average ASR over $66\%$. In~\cref{fig:det}, we showcase successful examples of object-selective hiding. The scale-dependent perturbations effectively obscure the targeted class when the perturbed image is resized to the predefined scaling factors used in transform-dependent optimization, demonstrating precise control over object concealment.

\noindent \textbf{Transform-selective hiding.} In this scenario, we hide all detectable objects when a certain level of image enhancement is applied to disclose sensitive information, and preserve detectability in unaltered or minimally enhanced images, enabling privacy protection and strategic information control. For instance, surveillance footage can obscure sensitive objects upon enhancement, preventing unauthorized recognition while maintaining visibility in unaltered conditions. Similarly, in satellite imaging, critical infrastructure can be concealed under specific enhancements, controlling detectability based on operational needs. 

We consider attacking zoom-in (\ie, scaling + centered cropping) with $S \sim [2.0, 2.5]$, blurring with $\sigma \sim [0.0, 0.7]$, and gamma correction with $\gamma \sim [0.5, 0.9]$ to simulate these enhancement scenarios. Outside of attack ranges, which we consider as safe ranges, we keep the detectability of objects and report accuracy.
\cref{tab:attack-detector-enhance} show that the attack is selectively triggered within the desired range of enhancement transform parameters, with overall ASR above $50\%$ over attack ranges and high ACC over safe ranges. Our attack consistently hides objects when different image enhancement transforms are heavily applied, while preserving detectability in unaltered or minimally enhanced images.
The examples in \cref{fig:det-enhance} show that objects are selectively hidden in predefined attack ranges in three enhance transforms, meanwhile, over safe range the detection results remain identical to original clean image. We provide animated visual examples of this attack scenario in the Supplemental Material.

\section{Conclusion}
\label{sec:conclusion}

In this work, we introduce transform-dependent adversarial attacks, highlighting vulnerability in deep networks from a novel perspective. Unlike conventional adversarial examples, which remain static in their effect, our findings reveal that a single perturbation can dynamically alter its adversarial impact based on image transformations. Through extensive experiments across various models and tasks, challenging blackbox and defensed setups, we demonstrate that these attacks enable precise, transformation-aware misclassifications, fundamentally challenging existing notions of adversarial robustness. 
Additionally, we motivate that such dynamic, transform-dependent property of adversarial examples can be used for image protection from potential sensitive information disclosure by enhancement transform.


{
    \small
    \bibliographystyle{ieeenat_fullname}
    \bibliography{ref}
}

\appendix
\newpage
\clearpage

\maketitlesupplementary

\newcommand{\beginsupplement}{%
        \setcounter{table}{0}
        \renewcommand{\thetable}{S\arabic{table}}%
        \setcounter{figure}{0}
        \renewcommand{\thefigure}{S\arabic{figure}}%
        \setcounter{section}{0}
        \renewcommand{\thesection}{S\arabic{section}}%
}
\makeatletter
\newcommand\refwithdefault[2]{%
  \@ifundefined{r@#1}{%
    #2%
  }{%
    \ref{#1}%
  }%
}
\makeatother

\section*{Summary}
In this supplementary material, we first present additional evaluations for transform-dependent attacks, in \cref{sec:additional}, including more challenging label selections, increasing number of transform-target pairs, and adaptation to alternative optimization methods. Then, we extend transform-dependent attacks to additional image transforms --- perspective and flipping, in \cref{sec:more-trans}, beyond those discussed in \cref{subsec:method-transform}. Finally, we provide an accuracy evaluation of the classifier models used in our main experiment (\cref{subsec:whitebox}), and additional visual examples of attacks on classifiers and detectors for reference.


\beginsupplement

\section{Additional evaluation} \label{sec:additional}
In this section, we first analyze the impact of increasing the number of transform-target pairs to stress test the capacity of adversarial perturbations in embedding transform-dependent targets in \cref{subsec:more-targets}. Next, we explore alternative optimization algorithms beyond PGD for generating transform-dependent attacks in \cref{subsec:diff-opt-algs}. Additionally, we evaluate attacks on more challenging label selection for attack optimization in \cref{subsec:challenging}.

\begin{figure*}[h]
    \centering

        \includegraphics[width=1.0\linewidth]{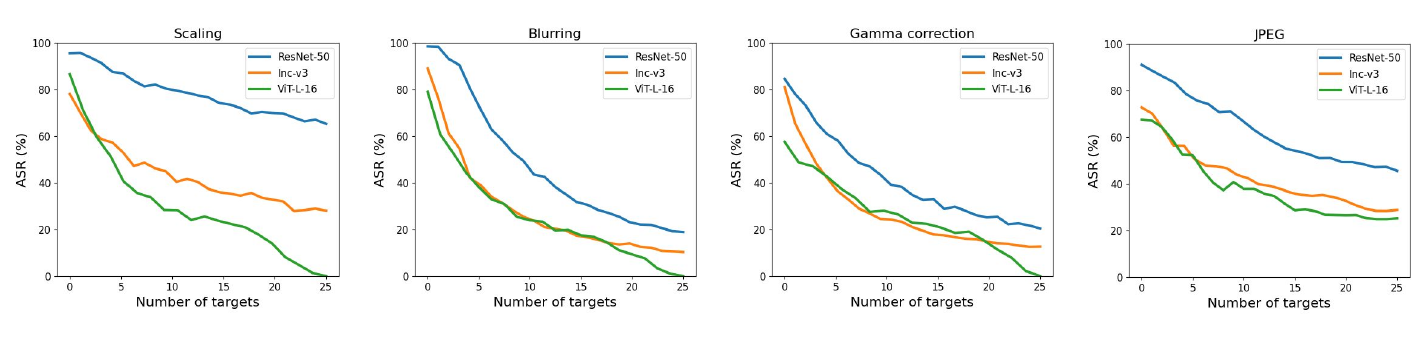}

    \caption{Average targeted ASR vs the number of embedded targets across scaling, blurring, gamma correction, and JPEG compression. As the number of embedded targets increases, the ASR drop indicates that the attack becomes more challenging. 
    }
    \label{fig:attack-trend}
\end{figure*}

\subsection{Increasing the Number of Targets}\label{subsec:more-targets}
Given what we introduced in this work that multiple targeted attacks can be embedded within a single perturbation through image transformations, a natural question arises: \textit{``How many transformation-dependent targets can be embedded in one attack perturbation?''}
To explore this, we conduct a study on three models sampled from \cref{subsec:whitebox}: ResNet-50, InceptionV3, and ViT-L-16, using scaling, blurring, gamma correction, and JPEG compression as transformations. We initialize the transform parameter sequence with ${0.5}$ for scaling, blurring, and gamma correction, and ${20}$ for JPEG compression. Additional $\theta_i$ values are iteratively appended using an adaptive step size until reaching a maximum of 25 samples. For each newly appended $\theta_i$, we assign a randomly sampled label from ImageNet label space as target $y_i^\star$.

The ASR trends in \cref{fig:attack-trend} show that as the number of samplings increases, the ASR drop rate varies across models and transformations. Scaling and JPEG compression demonstrate a higher capacity to embed multiple targeted attacks compared to blurring and gamma correction, this is consistent with the sensitivity indicate by loss landscape discussed in \cref{subsec:whitebox}. Among the tested models, ResNet-50 accommodates more transform-target pairs, whereas ViT-L-16 exhibits the smallest capacity.

\subsection{Adapt other optimization algorithms}\label{subsec:diff-opt-algs}
In our main paper, we primarily solve transform-dependent attacks using PGD~\cite{madry2017towards}, as discussed in \cref{subsec:method-prelim} for simplicity. However, in principle, these attacks can be generated with other optimization methods. Here, we evaluate the adaptability of transform-dependent attacks using commonly used methods, including FGSM~\cite{goodfellow2014explaining}, MIM~\cite{dong2018boosting}, and C\&W~\cite{carlini2017towards}, and compare their performance with PGD.

\cref{tab:attack-adapt} presents the average ASR for scale-dependent targeted attacks under the same settings as \cref{tab:cls-asr-range} in the main text. Among the tested methods, MIM, C\&W, and PGD --- being iterative approaches --- achieve high ASR, while the single-step FGSM proves insufficient for solving transform-dependent attacks.

\begin{table}[h]
\centering
\caption{Average scale-dependent ASRs (\%) $\uparrow$ with different attack optimization methods, perturbation budget $\varepsilon=8$. PGD offers overall better ASR among all optimization methods.}
\label{tab:attack-adapt}
\small
\begin{tabular}{c|cccc}
\hline
Attack method 
 & ResNet50 & VGG19 & Dense121 & Incv3 \\ \hline
FGSM~\cite{goodfellow2014explaining} & 0.13 & 0.13 & 0.07 & 0.10 \\
C\&W~\cite{carlini2017towards} & 85.90 & 90.73 & 90.77 & 60.50 \\
MIM~\cite{dong2018boosting} & 92.80 & 96.77 & 94.83 & \textbf{83.10} \\
PGD (\cref{tab:cls-asr-range})
& \textbf{96.06} & \textbf{98.83} & \textbf{97.10} & 82.08 \\ \hline
\end{tabular}


\end{table}

\subsection{More challenging attack targets}\label{subsec:challenging}
\begin{table*}[h]

\centering
\small
\caption{
ASR evaluation of transform-dependent attacks using 3 most least-likely labels as targets (challenging target selection). Higher value indicates better attack performance. The perturbation budget is $\varepsilon= 8$.
}
\label{tab:attack-hard}

\begin{tabular}{ccccccccc}
\hline
\multicolumn{1}{c|}{\multirow{2}{*}{\begin{tabular}[c]{@{}c@{}}Transform\\ parameter\end{tabular}}} & \multicolumn{8}{c}{Classifier model ASR(\%) $\uparrow$} \\
\multicolumn{1}{c|}{} & VGG19 & ResNet50 & Dense121 & Incv3 & Mobv2 & ViT-L-16 & ViT-L-32 & Swin-T \\ \hline
\multicolumn{1}{c|}{$S=0.5$} & 98.60 & 89.60 & 92.10 & 77.70 & 96.20 & 81.70 & 65.40 & 99.40 \\
\multicolumn{1}{c|}{$S=1.0$} & 100.0 & 99.70 & 99.70 & 91.00 & 100.0 & 95.40 & 85.80 & 100.0 \\
\multicolumn{1}{c|}{$S=1.5$} & 100.0 & 99.40 & 99.00 & 82.50 & 99.80 & 91.80 & 78.00 & 100.0 \\ \hline
\multicolumn{1}{c|}{Average} & 99.53 & 96.23 & 96.93 & 83.73 & 98.67 & 89.63 & 76.40 & 99.80 \\ \hline
\multicolumn{1}{c|}{$\sigma=0.5$} & 100.0 & 99.80 & 99.90 & 90.60 & 99.80 & 94.40 & 88.70 & 100.0 \\
\multicolumn{1}{c|}{$\sigma=1.5$} & 99.30 & 96.90 & 98.20 & 77.70 & 95.80 & 74.80 & 65.60 & 98.90 \\
\multicolumn{1}{c|}{$\sigma=3.0$} & 99.40 & 95.60 & 98.20 & 74.50 & 94.20 & 71.20 & 55.90 & 99.10 \\ \hline
\multicolumn{1}{c|}{Average} & 99.57 & 97.43 & 98.77 & 80.93 & 96.60 & 80.13 & 70.07 & 99.33 \\ \hline
\multicolumn{1}{c|}{$\gamma=0.5$} & 100.0 & 99.60 & 99.80 & 91.90 & 99.90 & 98.40 & 90.90 & 99.80 \\
\multicolumn{1}{c|}{$\gamma=1.0$} & 100.0 & 99.90 & 100.0 & 90.70 & 99.90 & 94.40 & 88.00 & 99.70 \\
\multicolumn{1}{c|}{$\gamma=2.0$} & 100.0 & 99.60 & 99.70 & 89.10 & 99.70 & 85.60 & 76.80 & 99.90 \\ \hline
\multicolumn{1}{c|}{Average} & 100.0 & 99.70 & 99.83 & 90.57 & 99.83 & 93.47 & 85.23 & 99.80 \\ \hline
\multicolumn{1}{c|}{$Q=20$} & 84.20 & 89.00 & 65.60 & 62.20 & 82.90 & 79.60 & 73.70 & 83.00 \\
\multicolumn{1}{c|}{$Q=50$} & 95.80 & 97.00 & 88.70 & 74.40 & 95.20 & 86.00 & 79.70 & 96.60 \\
\multicolumn{1}{c|}{$Q=80$} & 98.80 & 98.90 & 96.00 & 79.50 & 97.90 & 88.80 & 79.40 & 99.80 \\ \hline
\multicolumn{1}{c|}{Average} & 92.93 & 94.97 & 83.43 & 72.03 & 92.00 & 84.80 & 77.60 & 93.13 \\ \hline
\end{tabular}

\end{table*}

%
In \cref{subsec:whitebox} of the main text, we initially employed a random selection process to choose three distinct classes from the set of $1000$ ImageNet classes as our target labels, denoted as $\{y_i^\star\}_{i=1}^3$. Here, we maintain the consistent setup as \cref{sec:experiment} (i.e., datasets, models, hyperparameters), and opt to target label section as the \textbf{\textit{three least-likely labels}} extracted from the probability vectors converted from logits.

In this continuation, we evaluate the Adversarial Success Rate (ASR) under these modified attack settings, as presented in \cref{tab:attack-hard}. This evaluation demonstrates the efficacy of our attack formulation even when faced with the challenge of targeting the least-likely labels.

\section{More transformations}\label{sec:more-trans}
Conceptually, transform-dependent attack formulation applies to any differentiable and deterministic image transformation, as discussed in \cref{subsec:method-transform}. Here, we extend our approach to two additional geometric transforms as examples --- flipping and perspective transformation, to further demonstrate its flexibility.

\subsection{Flip-dependent attack}

In our examination of flip-dependent attacks, we explore three specific targeted scenarios designed to activate upon the network's receipt of images subjected to vertical flip, horizontal flip, and no flip (retaining their original orientation) as inputs. The findings, as detailed in ~\cref{tab:attack-flip}, reveal a notable efficacy of these flip-dependent attacks, achieving an ASR of 90\% across the majority of evaluated models.

\begin{table*}[h]

\centering
\small
\caption{Flip-dependent targeted attack success rate (ASR). ASRs are reported at each target flip version of image and average over all three flipping methods. Higher value indicates better attack performance. The perturbation budget is $\varepsilon= 8$. Below is the model classification accuracy (ACC) evaluation over flipped clean images without perturbation.}
\label{tab:attack-flip}

\begin{tabular}{ccccccccc}
\hline
\multicolumn{1}{c|}{\multirow{2}{*}{Flip method}} & \multicolumn{8}{c}{Classifier model}                                       \\
\multicolumn{1}{c|}{}                             & VGG19 & ResNet50 & Dense121 & Incv3 & Mobv2 & ViT-L-16 & ViT-L-32 & Swin-T \\ \hline
\multicolumn{9}{c}{ASR (\%) $\uparrow$}                                                                                        \\
\multicolumn{1}{c|}{None}                         & 100.0 & 99.70    & 99.90    & 92.10 & 100.0 & 93.40    & 85.50    & 100.0  \\
\multicolumn{1}{c|}{Horizontal}                   & 100.0 & 99.90    & 99.80    & 91.70 & 100.0 & 92.80    & 87.20    & 100.0  \\
\multicolumn{1}{c|}{Vertical}                     & 100.0 & 99.80    & 99.90    & 97.70 & 100.0 & 96.10    & 91.40    & 100.0  \\ \hline
\multicolumn{1}{c|}{Average}                      & 100.0 & 99.80    & 99.87    & 93.83 & 100.0 & 94.10    & 88.03    & 100.0  \\ \hline
\multicolumn{9}{c}{ACC (\%) $\uparrow$}                                                                                        \\
\multicolumn{1}{c|}{None}                         & 100.0 & 100.0    & 100.0    & 100.0 & 87.60 & 100.0    & 100.0    & 96.00  \\
\multicolumn{1}{c|}{Horizontal}                   & 93.20 & 94.20    & 96.20    & 80.60 & 87.90 & 98.30    & 96.40    & 96.10  \\
\multicolumn{1}{c|}{Vertical}                     & 55.00 & 57.40    & 62.00    & 39.30 & 53.10 & 72.80    & 54.80    & 78.00  \\ \hline
\multicolumn{1}{c|}{Average}                      & 82.73 & 83.87    & 86.07    & 73.30 & 76.20 & 90.37    & 83.73    & 90.03  \\ \hline
\end{tabular}

\end{table*}
\begin{table*}[h]

\centering
\small
\caption{Perspective-dependent targeted attack success rate (ASR). 
ASRs are reported at each target flip version of image and average over all three flipping methods. Higher value indicates better attack performance.
The perturbation budget is $\varepsilon= 8$.
Below is the model classification accuracy (ACC) evaluation over perspective transformed clean images without perturbation.
}
\label{tab:attack-perspe}

\begin{tabular}{ccccccccc}
\hline
\multicolumn{1}{c|}{\multirow{2}{*}{Perspective}} & \multicolumn{8}{c}{Classifier model}                                       \\
\multicolumn{1}{c|}{}                             & VGG19 & ResNet50 & Dense121 & Incv3 & Mobv2 & ViT-L-16 & ViT-L-32 & Swin-T \\ \hline
\multicolumn{9}{c}{ASR (\%) $\uparrow$}                                                                                        \\
\multicolumn{1}{c|}{1}                            & 100.0 & 99.70    & 100.0    & 92.90 & 100.0 & 96.70    & 91.30    & 100.0  \\
\multicolumn{1}{c|}{2}                            & 99.80 & 99.20    & 99.40    & 87.70 & 100.0 & 83.30    & 71.30    & 99.90  \\
\multicolumn{1}{c|}{3}                            & 99.90 & 99.50    & 99.10    & 92.50 & 100.0 & 83.80    & 69.20    & 100.0  \\ \hline
\multicolumn{1}{c|}{Average}                      & 99.90 & 99.47    & 99.50    & 91.03 & 100.0 & 87.93    & 77.27    & 99.97  \\ \hline
\multicolumn{9}{c}{ACC (\%) $\uparrow$}                                                                                        \\
\multicolumn{1}{c|}{1}                            & 90.80 & 93.20    & 94.50    & 79.90 & 87.90 & 92.70    & 89.90    & 96.10  \\
\multicolumn{1}{c|}{2}                            & 72.10 & 72.10    & 73.50    & 53.00 & 65.30 & 87.80    & 79.30    & 91.30  \\
\multicolumn{1}{c|}{3}                            & 73.60 & 75.40    & 74.10    & 57.90 & 63.50 & 87.50    & 78.20    & 92.10  \\ \hline
\multicolumn{1}{c|}{Average}                      & 78.83 & 80.23    & 80.70    & 63.60 & 72.23 & 89.33    & 82.47    & 93.17  \\ \hline
\end{tabular}

\end{table*}

\subsection{Perspective-dependent attack}

Here, we introduce perspective-dependent attacks to mimic the variability encountered when taking photos from different angles. We categorize these variations into three predefined perspectives: viewing the subject from the front, from above, and from below, labeled as perspectives 1, 2, and 3, respectively. Specifically, perspective 1 maintains the image in its original state, illustrating a front-facing viewpoint. Perspective 2 simulates a downward view by transforming the image axis from \{(0,0), (223,0)\} to \{(56,56), (168,56)\}, and perspective 3 simulates an upward view by altering the image axis from \{(0,223), (223,223)\} to \{(56,168), (168,168)\}.

The results, as noted in ~\cref{tab:attack-perspe}, demonstrate the effectiveness of these perspective-dependent attacks, with targeted strategies achieving an overall ASR of over 90\% when images are presented from these varied perspectives.

\begin{table*}[h]

\centering
\small
\caption{
Clean accuracy (ACC) evaluation over selected classification models.
Higher value indicating lower classification error introduced by the image transformation.
}
\label{tab:cls-acc}

\begin{tabular}{ccccccccc}
\hline
\multicolumn{1}{c|}{\multirow{2}{*}{\begin{tabular}[c]{@{}c@{}}Transform\\ parameter\end{tabular}}} & \multicolumn{8}{c}{Classifier model} \\
\multicolumn{1}{c|}{}             & VGG19 & ResNet50 & Dense121 & Incv3 & Mobv2 & ViT-L-16 & ViT-L-32 & Swin-T \\ \hline
\multicolumn{1}{c|}{$S=0.5$}      & 66.00 & 69.20    & 62.20    & 29.30 & 56.10 & 89.10    & 83.00    & 78.30  \\
\multicolumn{1}{c|}{$S=1.0$}      & 100.0 & 100.0    & 100.0    & 100.0 & 87.90 & 100.0    & 100.0    & 96.10  \\
\multicolumn{1}{c|}{$S=1.5$}      & 87.80 & 90.40    & 92.00    & 79.80 & 83.60 & 97.80    & 96.40    & 92.50  \\ \hline
\multicolumn{1}{c|}{Average}      & 84.60 & 86.53    & 84.73    & 69.70 & 75.87 & 95.63    & 93.13    & 88.97  \\ \hline
\multicolumn{1}{c|}{$\sigma=0.5$} & 94.20 & 95.80    & 96.90    & 91.50 & 88.10 & 98.30    & 96.90    & 94.70  \\
\multicolumn{1}{c|}{$\sigma=1.5$} & 71.00 & 76.50    & 78.30    & 67.00 & 60.30 & 88.20    & 81.90    & 81.10  \\
\multicolumn{1}{c|}{$\sigma=3.0$} & 64.60 & 74.10    & 75.80    & 61.10 & 58.50 & 85.90    & 78.40    & 79.10  \\ \hline
\multicolumn{1}{c|}{Average}      & 76.60 & 82.13    & 83.67    & 73.20 & 68.97 & 90.80    & 85.73    & 84.97  \\ \hline
\multicolumn{1}{c|}{$\gamma=0.5$} & 91.90 & 92.30    & 95.80    & 85.00 & 82.80 & 94.40    & 92.70    & 93.70  \\
\multicolumn{1}{c|}{$\gamma=1.0$} & 100.0 & 100.0    & 100.0    & 100.0 & 87.90 & 100.0    & 100.0    & 96.10  \\
\multicolumn{1}{c|}{$\gamma=2.0$} & 90.90 & 90.70    & 92.80    & 80.60 & 82.40 & 91.80    & 88.60    & 95.00  \\ \hline
\multicolumn{1}{c|}{Average}      & 94.27 & 94.33    & 96.20    & 88.53 & 84.37 & 95.40    & 93.77    & 94.93  \\ \hline
\multicolumn{1}{c|}{$Q=20$}       & 71.30 & 78.20    & 84.10    & 68.90 & 69.80 & 81.70    & 83.30    & 65.90  \\
\multicolumn{1}{c|}{$Q=50$}       & 82.10 & 86.10    & 89.60    & 75.50 & 77.10 & 87.50    & 88.00    & 82.30  \\
\multicolumn{1}{c|}{$Q=80$}       & 87.30 & 89.50    & 92.10    & 78.00 & 84.80 & 91.20    & 89.40    & 89.00  \\ \hline
\multicolumn{1}{c|}{Average}      & 80.23 & 84.60    & 88.60    & 74.13 & 77.23 & 86.80    & 86.90    & 79.07  \\ \hline
\end{tabular}

\end{table*}

\section{Accuracy evaluations} 
In \cref{subsec:whitebox} of our main paper, we presented targeted attacks designed to exploit vulnerabilities specific to scaling, blurring, gamma correction, and JPEG compression. To distinguish the adversarial effects from mere consequences of image transformations, we evaluate Attack Success Rate (ASR) across a diverse set of models: \{VGG-19-BN, ResNet50, DenseNet-121, InceptionV3, ViT-L-16, ViT-L-32, Swin-T\}.

To further investigate model sensitivity to transformations, \cref{tab:cls-acc} reports classification accuracy on clean images subjected to the same transformations used in attack generation. While most models maintain high accuracy (e.g., over 80\%), one architecture, InceptionV3, exhibits notable sensitivity, with accuracy dropping to 69.70\%. We mitigate this influence in our experiment by experimenting on sufficiently diverse set of models, and image transformations, and dataset contains larger number of instances, following similar principle to prior adversarial works that utilize image transformation for adversarial examples generation discussed in \cref{subsec:related-trans-attack}.

\newpage

\section{Memory and computation resources. } We used a single {NVIDIA RTX 2080Ti (12 GB)} for all the experiments. Average times for generating $\{3,5,10\}$ target attacks are $\{2.61, 4.29, 8.67\}$ sec/image.

\section{More visual examples}\label{sec:more-visual}
\begin{figure*}[t]
\centering
%
\begin{subfigure}[c]{0.85\linewidth}
    \centering
    \includegraphics[width=1.0\textwidth]{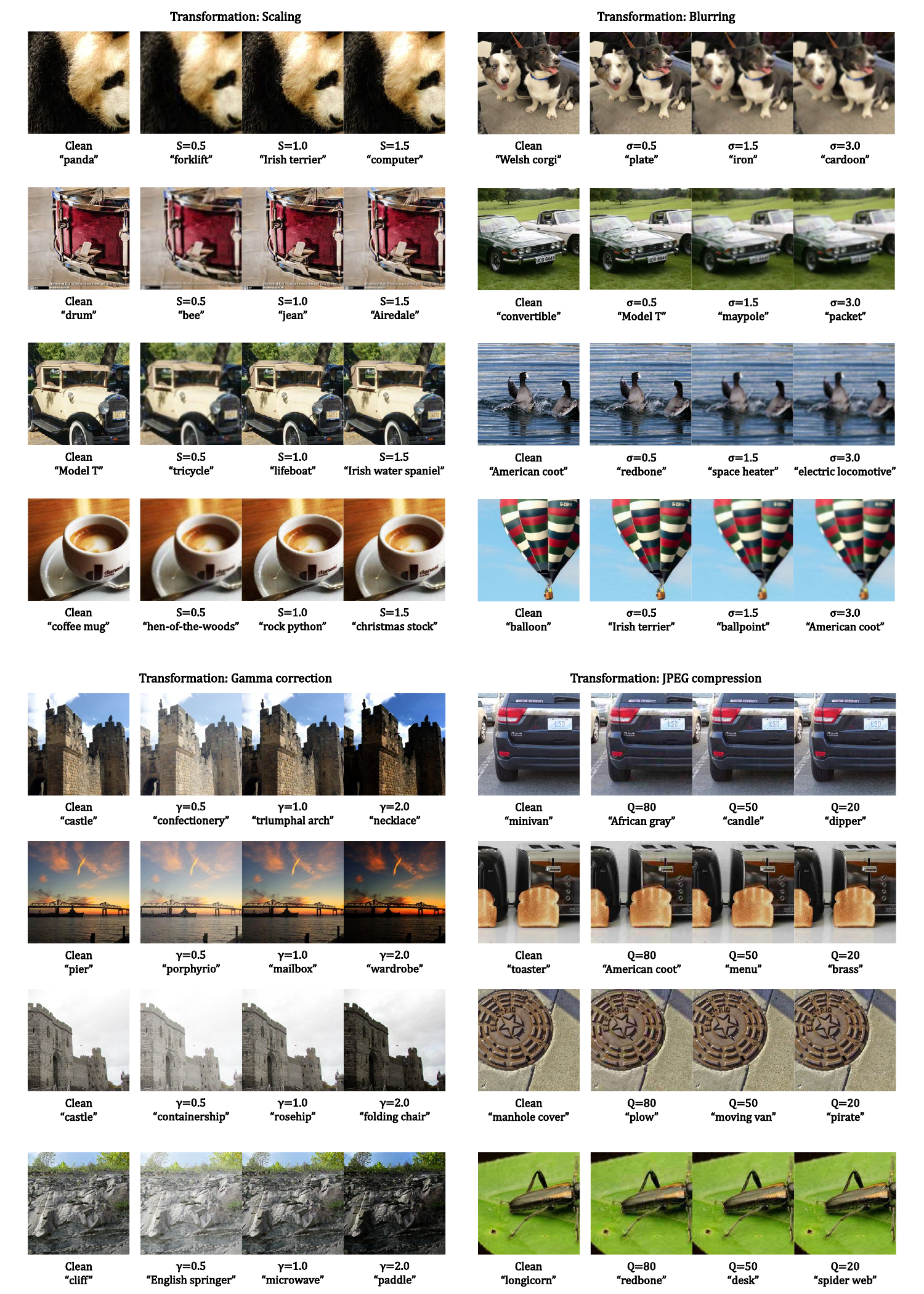}
\end{subfigure}
\caption{Visualize examples for transform-dependent attacks against image classifiers. In this figure, we show visual effects of clean image and the perturbed images transformed with different parameters.}
\label{fig:cls-supp-1}
\end{figure*}

\begin{figure*}[t]
\centering
%
\begin{subfigure}[c]{0.85\linewidth}
    \centering
    \includegraphics[width=1.0\textwidth]{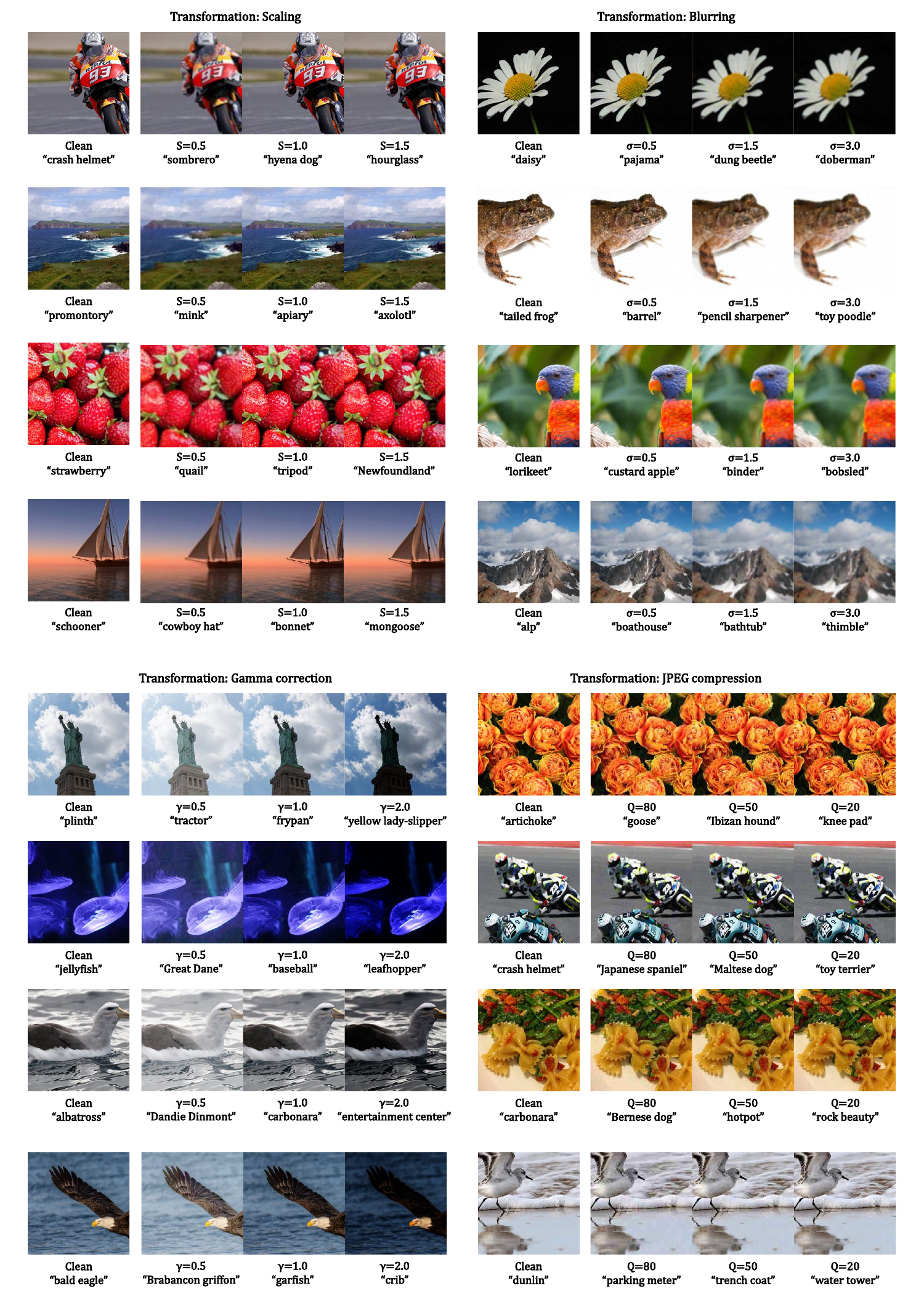}
\end{subfigure}
\caption{Visualize examples for transform-dependent attacks against image classifiers. In this figure, we show visual effects of successful attacks under different image transformations with different transform parameters.}
\label{fig:cls-supp-2}
\end{figure*}

\begin{figure*}[t]
\centering
%
\begin{subfigure}[c]{0.95\linewidth}
    \centering
    \includegraphics[width=1.0\textwidth]{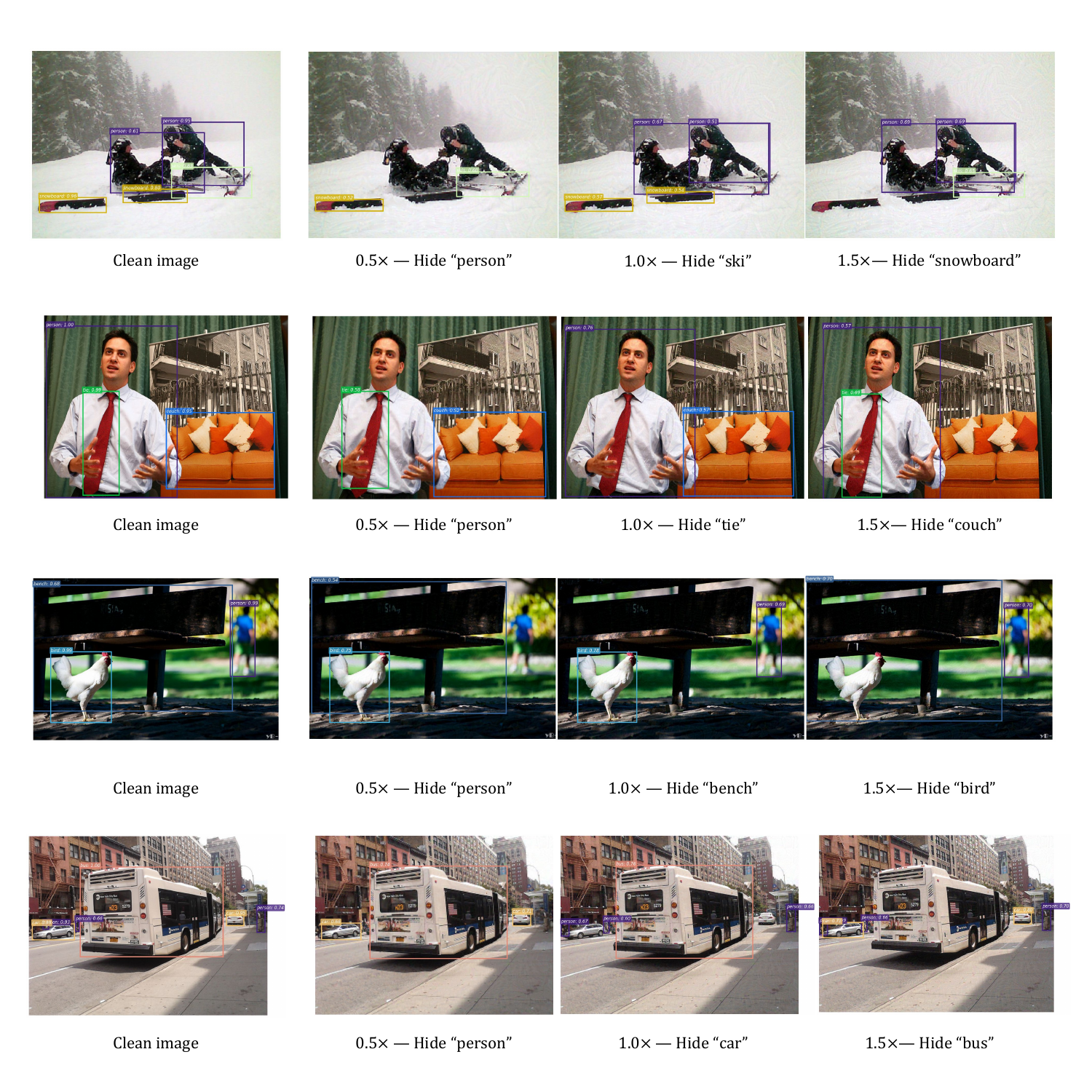}
\end{subfigure}
\caption{Visualize examples of scale-dependent selective hiding attacks against object detection model \texttt{FCOS}. From top to button, \texttt{ImageIDs: 000000142790, 000000170099, 000000197870, 000000338625}. Note, images labeled as $0.5, 1.5\times$ are in resolutions different from the original image, and they are resized to the same size for better display.}
\label{fig:det-supp-1}
\end{figure*}

\begin{figure*}[t]
\centering
%
\begin{subfigure}[c]{0.95\linewidth}
    \centering
    \includegraphics[width=1.0\textwidth]{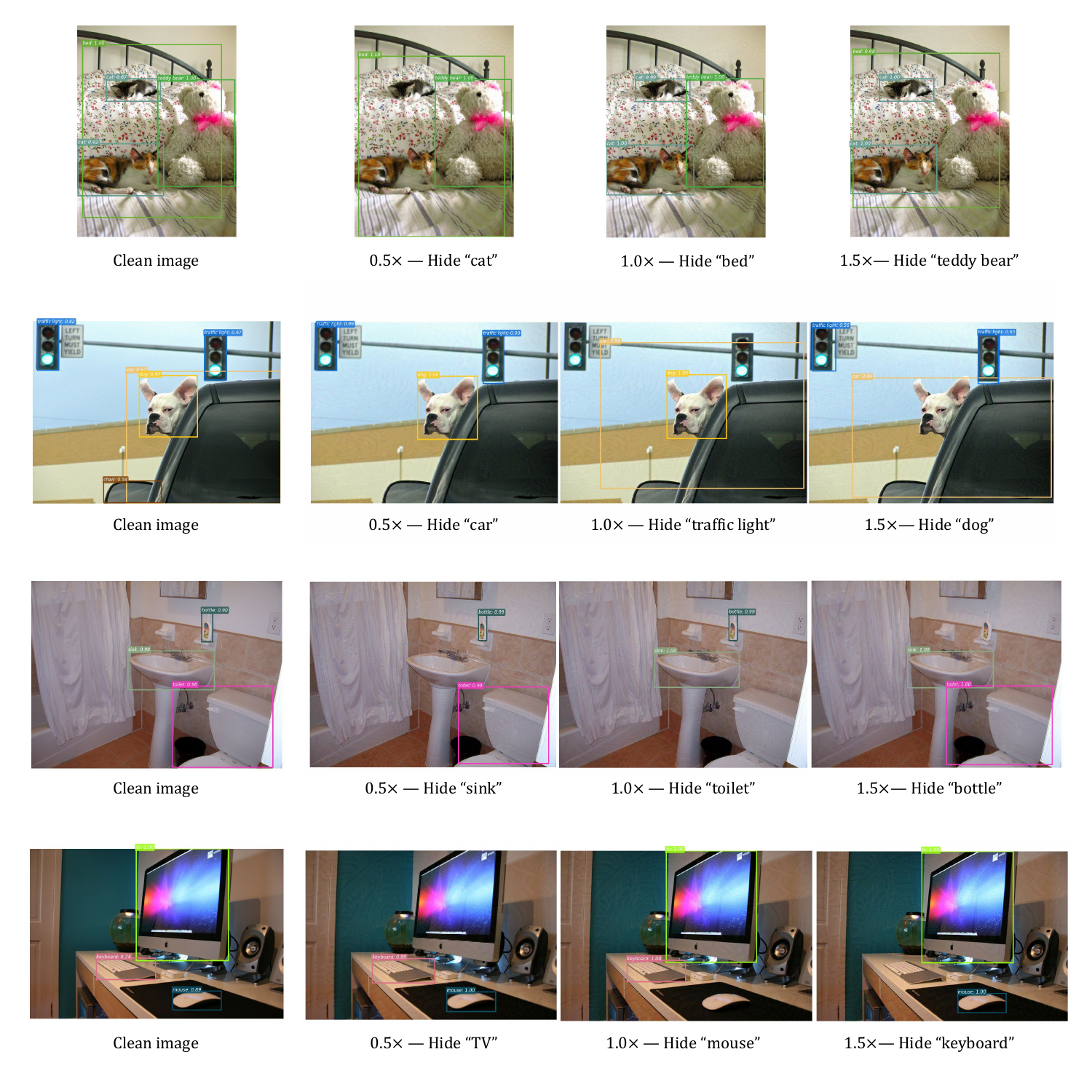}
\end{subfigure}
\caption{Visualize examples of scale-dependent selective hiding attacks against object detection model \texttt{YOLOv3}. From top to button, \texttt{ImageIDs: 000000478393, 000000076417, 000000167898, 000000186282}. Note, images labeled as $0.5, 1.5\times$ are in resolutions different from the original image, and they are resized to the same size for better display.}
\label{fig:det-supp-2}
\end{figure*}

\begin{figure*}[t]
\centering
%
\begin{subfigure}[c]{0.95\linewidth}
    \centering
    \includegraphics[width=1.0\textwidth]{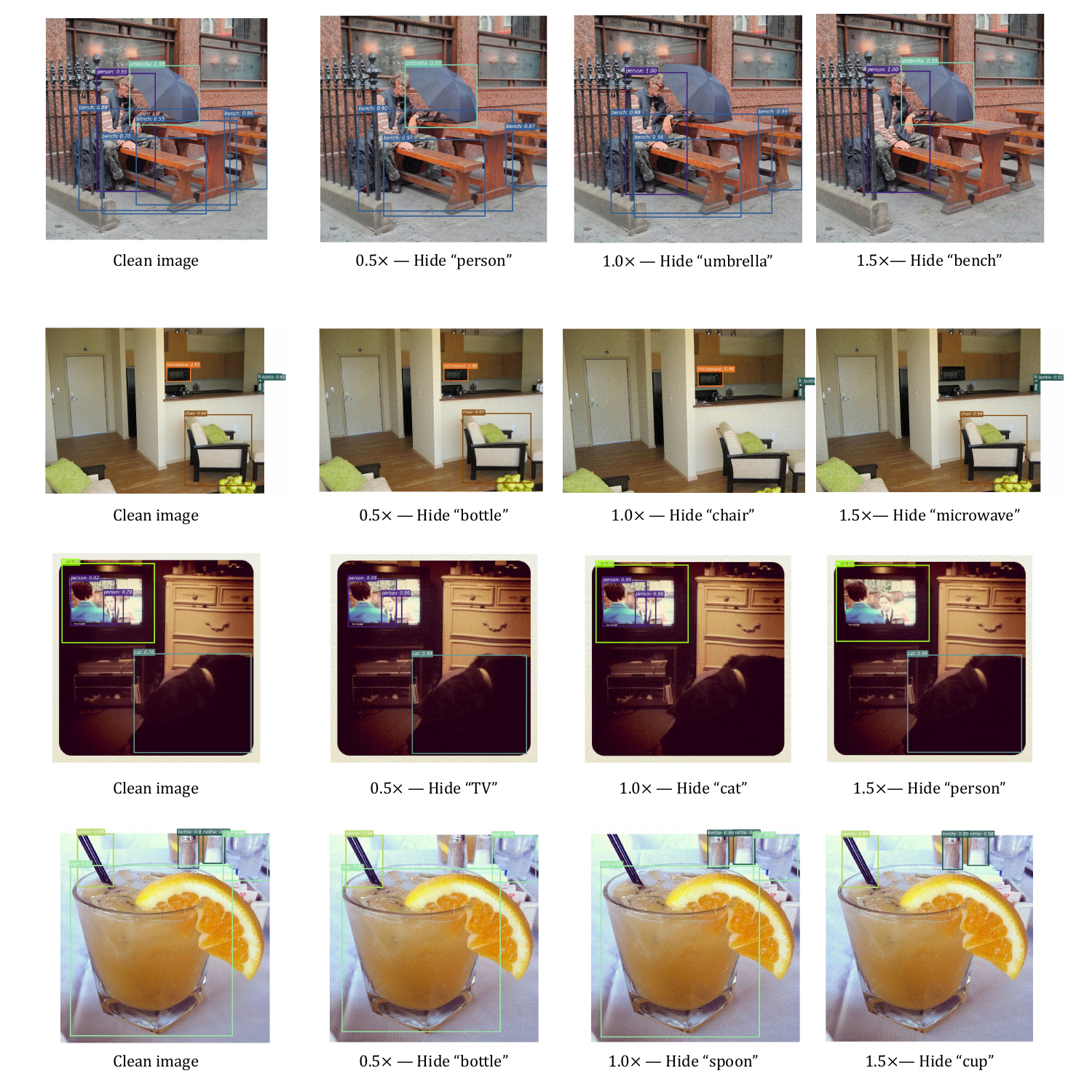}
\end{subfigure}
\caption{Visualize examples of scale-dependent selective hiding attacks against object detection model \texttt{Faster R-CNN}. From top to button, \texttt{ImageIDs: 000000455157, 000000488075, 000000169076, 000000463283}. Note, images labeled as $0.5, 1.5\times$ are in resolutions different from the original image, and they are resized to the same size for better display.}
\label{fig:det-supp-3}
\end{figure*}

For enhanced qualitative evaluation, we offer additional visual examples showcasing successful transform-dependent adversarial instances against image classification models in~\cref{fig:cls-supp-1} and \cref{fig:cls-supp-2}. Moreover, we present further examples of object-selective hiding attacks against object detection models in~\cref{fig:det-supp-1}, \cref{fig:det-supp-2}, and~\cref{fig:det-supp-3}. 
For transform-selective hiding attacks against detectors, which we demonstrate as a defense mechanism against image enhancement (\cref{fig:det-enhance} in the main text), we provide additional visualizations in the supplementary material. Specifically, we include animations in \texttt{.gif} format to illustrate how object detectability changes dynamically under different enhancement transformations.

\noindent\textbf{Attacks against classifiers.} The examples in~\cref{fig:cls-supp-1} and \cref{fig:cls-supp-2} illustrate that with imperceptible noise perturbation, an image can be misclassified as multiple target labels when subjected to specific image transformations.

\noindent\textbf{Attacks against detectors.} In the scenario of object-selective hiding attacks, consider the images in the first row of \cref{fig:det-supp-1} as an illustration. It demonstrates that in the clean image, objects labeled as three distinct categories (\textit{person, ski}, and \textit{snowboard}) are detected. However, upon adding scale-dependent perturbations, objects labeled as one of these categories become hidden in three differently scaled versions of perturbed images, as depicted in the titles: \textit{$0.5\times$ --- Hide "person"}, \textit{$1.0\times$ --- Hide "ski"}, and \textit{$1.5\times$ --- Hide "snowboard"}.

In the scenario of transform-selective hiding attacks, we simulate image enhancement processes—zoom-in, deblur, and gamma correction—applied to raw images initially presented as distant scenes, blurry images, or low-light conditions. Some illustrative examples are publicly available at this \href{https://drive.google.com/drive/folders/1DPpqWRSFHWYe8FRz3HkFZsCm9gqSTpOv?usp=sharing}{\underline{Google Drive}}.
The animations illustrate the effectiveness of our transform-dependent perturbations: objects remain detectable in unaltered images but become concealed when the images undergo enhancement. This demonstrates the perturbation’s ability to selectively obscure objects under specific transformations while preserving detectability close to the original, non-perturbed state.


\end{document}